\documentclass[11pt, a4paper]{article}

\usepackage[utf8]{inputenc} 
\usepackage[T1]{fontenc}   
\usepackage{amsmath}        
\usepackage{graphicx}       
\usepackage{booktabs}       
\usepackage[margin=1in]{geometry} 
\usepackage{hyperref}       
\usepackage{parskip}        
\usepackage{caption}        
\usepackage{url}            
\usepackage{xurl}           
\usepackage{float}          
\usepackage{microtype}      
\usepackage{placeins}       
\usepackage{tabularx}

\title{Predicting High NCAP Safety Ratings: An Analysis of Vehicle Characteristics and ADAS Features using Machine Learning}

\author{Raunak Kunwar$^1$ \and Dr. Aera Kim LeBoulluec$^2$}

\date{}

\begin{document}

\begin{titlepage}
    \maketitle
    \vspace{2em}
    \noindent $^1$\textit{University of Texas at Arlington, 701 S. Nedderman Drive, Arlington, TX 76019, USA, raunakkunwar999@gmail.com} \\ 
    \noindent $^2$\textit{University of Texas at Arlington, 701 S. Nedderman Drive, Arlington, TX 76019, USA, aeral@uta.edu} \\
    \vspace{1em}
    
    \clearpage
\end{titlepage}

% --- ABSTRACT ---
\begin{abstract}
Vehicle safety assessment is crucial for consumer information and regulatory oversight. The New Car Assessment Program (NCAP) assigns standardized safety ratings, which traditionally emphasize passive safety measures but now include active safety technologies such as Advanced Driver-Assistance Systems (ADAS). It is crucial to understand how these various systems interact empirically. This study explores whether particular ADAS features like Forward Collision Warning, Lane Departure Warning, Crash Imminent Braking, and Blind Spot Detection, together with established vehicle attributes (e.g., Curb Weight, Model Year, Vehicle Type, Drive Train), can reliably predict a vehicle's likelihood of earning the highest (5-star) overall NCAP rating. Using a publicly available dataset derived from
NCAP reports that contain approximately 5,128 vehicle variants spanning model years 2011–2025, we compared four different machine learning models: logistic regression, random forest, gradient boosting, and support vector classifier (SVC) using a 5-fold stratified cross-validation approach. The two best-performing algorithms (random forest and gradient boost) were hyperparameter optimized using RandomizedSearchCV. Analysis of feature importance showed that basic vehicle characteristics, specifically curb weight and model year, dominated predictive capability, contributing more than 55\% of the feature relevance of the Random Forest model. However, the inclusion of ADAS features also provided meaningful predictive contributions. The optimized Random Forest model achieved robust results on a held-out test set, with an accuracy of 89.18\% and a ROC AUC of 0.9586. This research reveals the use of machine learning to analyze large-scale NCAP data and highlights the combined predictive importance of both established vehicle parameters and modern ADAS features to achieve top safety ratings.

\end{abstract}

% --- KEYWORDS ---
\noindent\textbf{Keywords:} NCAP; Vehicle Safety Ratings; ADAS; Passive Safety; Machine Learning; Predictive Modeling. 

% --- SECTIONS (Numbered) ---
\section{Introduction}

Vehicle safety continues to be a critical priority for manufacturers, regulatory bodies, and consumers worldwide. Programs like the New Car Assessment Program (NCAP), managed by organizations such as the National Highway Traffic Safety Administration (NHTSA) in the United States, provide standardized safety ratings derived primarily from crashworthiness assessments (passive safety) and rollover resistance evaluations \cite{nhtsa_ratings, nhtsa_resources}.  These programs originated in the late 1970s and have evolved significantly over time \cite{wiki_ncap}. Ratings serve as vital information for consumer purchasing decisions and drive manufacturers to improve vehicle safety features.  Historically, these assessments have focused primarily on passive safety components, such as airbags, seatbelts, and structural designs that aim to protect vehicle occupants during collisions \cite{elvik_2019, eckell_activepassive}.

In recent decades, the automotive landscape has been transformed by the rapid development and deployment of Advanced Driver-Assistance Systems (ADAS). These active safety systems, including features such as Forward Collision Warning (FCW), Automatic Emergency Braking (AEB), Lane Departure Warning (LDW), and Blind Spot Detection (BSD), aim to prevent crashes or mitigate their severity by assisting the driver or automatically intervening \cite{elvik_2019, parts_2025, caradas_activepassive}. Reflecting this technological shift, NCAP protocols are evolving to incorporate the assessment and rating of these active systems alongside traditional passive safety evaluations \cite{nhtsa_ratings, siemens_truckncap}. While numerous studies have demonstrated the effectiveness of specific ADAS in reducing certain types of crashes \cite{parts_2025}, understanding the complex interplay between these newer active systems and established passive safety performance using large-scale empirical data remains a significant challenge \cite{hu_2024, toyota_csrc, ircobi_integrated_assess}. Often, analyses rely on simulation, controlled tests, or real-world crash data that may not directly correspond to standardized NCAP testing conditions.

This study aims to bridge this gap by leveraging a large, publicly available dataset derived from NCAP test results to explore the predictive relationship between ADAS feature availability and overall passive safety outcomes, as represented by the NCAP 5-Star Safety Ratings. Using machine learning techniques \cite{kim_2024, researchgate_ml_rating}, which are increasingly applied in vehicle safety analysis, we seek to determine the extent to which the presence of specific ADAS features (FCW, LDW, Crash Imminent Braking, BSD) can predict the likelihood of a vehicle achieving a high (5-star) overall NCAP safety rating, while controlling for fundamental vehicle characteristics (Model Year, Vehicle Type, Drive Train, Curb Weight). In doing so, we also compare the predictive performance of four distinct machine learning algorithms (Logistic Regression, Random Forest, Gradient Boosting, Support Vector Classifier) and assess the relative importance of ADAS features compared to traditional vehicle characteristics for this prediction task.

By applying this comparative machine learning approach to a comprehensive empirical NCAP dataset, this study seeks to provide quantitative insights into the factors associated with top safety performance in modern vehicles. The findings aim to contribute to a better understanding of the combined predictive value of passive and active safety elements within the standardized NCAP framework. The remainder of the paper outlines the methodology adopted, presents detailed modeling results, discusses key implications and limitations, and concludes with final observations.

\section{Methodology}
This section details the data source, preparation steps, feature selection, modeling techniques, and evaluation strategy used in this study.

\subsection{Data Source and Preparation}
The analysis used a publicly available dataset compiled from New Car Assessment Program (NCAP) information, implicitly sourced from the National Highway Traffic Safety Administration (NHTSA) via the file \texttt{Safercar\_data.csv}. This dataset contains safety ratings, vehicle specifications, and information on safety features for numerous vehicle makes, models, and configurations, primarily from model years 2011 to 2025. Initial data preparation involved loading the dataset using the pandas library in Python and systematically replacing string representations of missing values (e.g., `nan') with NumPy's standard Not a Number (NaN) representation. A critical step involved handling the primary outcome variable, \texttt{OVERALL\_STARS}. Records where this value was missing or could not be unambiguously converted to a numeric representation were removed from the dataset to ensure a valid target variable for supervised learning. This resulted in a final analytical dataset comprising approximately 5,128 unique vehicle variant entries.

\subsection{Target Variable Definition}
The primary research objective was to predict whether a vehicle achieved the highest possible overall NCAP safety rating. To facilitate a binary classification task, the numeric \texttt{OVERALL\_STARS} value was transformed into a binary target variable named \texttt{high\_overall\_rating}. This variable was assigned a value of 1 for vehicles with a 5-star rating and a value of 0 for vehicles with ratings from 1 to 4 stars. The resulting distribution showed a moderate class imbalance, with approximately 63\% of the instances belonging to the positive class (5 stars).

\subsection{Feature Selection and Preprocessing}
Predictor variables (features) were selected to align with the research objectives, encompassing both ADAS availability and fundamental vehicle characteristics known to influence safety performance. The ADAS features included categorical variables indicating the presence/availability of FCW, LDW, Crash Imminent Braking, and BSD. Vehicle characteristics included numeric features (MODEL\_YR, CURB\_WEIGHT) and categorical features (VEHICLE\_TYPE, DRIVE\_TRAIN).

Extensive preprocessing was applied to these selected features using scikit-learn, integrated into a unified pipeline using \texttt{ColumnTransformer} for differential treatment of column types. First, standardization of categorical features was performed using custom functions to map varied entries within the ADAS columns (corresponding to FCW, LDW, Crash Imminent Braking, and BSD) into consistent categories (`Standard', `Optional', `None') and to map DRIVE\_TRAIN entries to standardized categories (e.g., `AWD', `FWD', `RWD', `4WD', `2WD', `Unknown'). Second, missing values were handled: remaining missing values in standardized ADAS columns were mapped to `None', while missing values in VEHICLE\_TYPE and DRIVE\_TRAIN were imputed with the string `Unknown'. Missing numeric values in the CURB\_WEIGHT column were imputed using the median value calculated from the training data partition to prevent data leakage (MODEL\_YR had no missing values in the modeling dataset). Third, categorical features were encoded into a numerical format using \texttt{OneHotEncoder}, which creates binary columns for each category and handles potential unknown categories in the test set. Fourth, numeric features (MODEL\_YR, CURB\_WEIGHT) were scaled using \texttt{StandardScaler} to have zero mean and unit variance, preventing features with larger values from disproportionately influencing model training. The final feature set after these preprocessing steps consisted of 24 features.

\subsection{Modeling and Evaluation}
The core of the methodology involved training, comparing, and evaluating four distinct machine learning classification models, encapsulated within scikit-learn \texttt{Pipeline} objects to ensure preprocessing steps were appropriately fitted only on training data during cross-validation and final model training, thus preventing data leakage. The models chosen were:

\subsubsection{Logistic Regression}
A linear model is used as a baseline. It models the probability of the binary outcome ($y=1$) using the sigmoid function applied to a linear combination of the input features ($\mathbf{X}$), typically represented as $P(y=1|\mathbf{X}) = \sigma(\beta_0 + \boldsymbol{\beta} \cdot \mathbf{X}) = 1 / (1 + \exp(-(\beta_0 + \boldsymbol{\beta} \cdot \mathbf{X})))$. The coefficients ($\beta_0, \boldsymbol{\beta}$) are learned by maximizing the likelihood of observing the training data, often using optimization algorithms like gradient descent.

\subsubsection{Random Forest Classifier}
An ensemble method based on constructing a multitude of decision trees ($T$) during training. It operates by building each tree on a bootstrapped sample (bagging) of the training data and considering only a random subset of features at each split to decorrelate the trees. For classification, the final prediction for an input $\mathbf{x}$ is determined by a majority vote among all individual trees ($h_t(\mathbf{x})$): $\hat{y} = \text{majority vote}\{h_1(\mathbf{x}), h_2(\mathbf{x}), ..., h_T(\mathbf{x})\}$. This approach makes it robust to overfitting and effective at capturing non-linearities.

\subsubsection{Gradient Boosting Classifier}
Another powerful ensemble technique that builds models (typically decision trees) sequentially. Each new model $h_m(\mathbf{x})$ attempts to correct the errors made by the ensemble of previous models $F_{m-1}(\mathbf{x})$ by fitting to the negative gradient (pseudo-residuals) of a chosen loss function $L(y, F)$ with respect to the previous prediction. The final prediction is an additive combination of the weak learners, updated iteratively: $F_m(\mathbf{x}) = F_{m-1}(\mathbf{x}) + \nu \cdot h_m(\mathbf{x})$, where $\nu$ is the learning rate. This iterative process often yields high predictive accuracy.

\subsubsection{Support Vector Classifier (SVC)}
A kernel-based method effective in high-dimensional spaces. For linearly separable data, SVC aims to find the optimal hyperplane $\mathbf{w} \cdot \mathbf{x} - b = 0$ that maximizes the margin (distance) between the data points of the different classes ($y_i \in \{-1, 1\}$). The core idea involves solving a convex optimization problem, typically formulated as minimizing the norm of the weight vector $\mathbf{w}$ subject to constraints ensuring correct classification with a defined margin: $\min_{\mathbf{w}, b} \frac{1}{2} ||\mathbf{w}||^2$ subject to $y_i (\mathbf{w} \cdot \mathbf{x}_i - b) \geq 1$ for all training points $(\mathbf{x}_i, y_i)$. For non-linear separation, SVC uses kernel functions $K(\mathbf{x}_i, \mathbf{x}_j)$ (like RBF, polynomial) to implicitly map data into higher-dimensional feature spaces where a linear separation might be possible.

The evaluation strategy includes several stages. First, the dataset (features X, target y) was divided into a training set (80\%) and a held-out test set (20\%) using stratification based on the \texttt{high\_overall\_rating} target variable to ensure similar class distributions. Second, an initial model comparison was conducted using the default hyperparameters for each of the four models. Performance was compared using 5-fold stratified cross-validation (via \texttt{StratifiedKFold}) executed on the training set. The assessment metrics included Accuracy, Precision (macro average), Recall (macro average), F1-score (macro average), and Area Under the Receiver Operating Characteristic Curve (ROC AUC), with macro averaging chosen to give equal weight to each class. Third, feature importance analysis was performed by extracting the \texttt{feature\_importances\_} attribute from the Random Forest and Gradient Boosting models after training them on the full training dataset. Fourth, hyperparameter tuning was conducted for the two best-performing models from the initial comparison (Random Forest and Gradient Boosting). \texttt{RandomizedSearchCV} was employed with 20 iterations per model and 5-fold stratified cross-validation on the training set, optimizing hyperparameters to maximize the mean cross-validation ROC AUC score. Finally, the best estimator (pipeline with optimized hyperparameters) identified by \texttt{RandomizedSearchCV} for both models was evaluated on the previously untouched test set using the same metrics as the cross-validation phase, added with detailed classification reports and confusion matrices for a comprehensive assessment of generalization performance.

\section{Results}
This section presents the outcomes of the machine learning models developed to predict high (5-star) overall NCAP safety ratings based on vehicle characteristics and ADAS features. The results are presented following the evaluation strategy outlined in the Methodology.

\subsection{Initial Model Performance Comparison}
The performance of the four machine learning models (Logistic Regression, Random Forest, Gradient Boosting, Support Vector Classifier) using their default hyperparameters was evaluated using 5-fold stratified cross-validation on the training dataset. Table \ref{tab:cv_results} summarizes the average performance across the folds for each metric.

% --- TABLE 1 ---
\begin{table}[htbp]
\centering
\caption{Cross-Validation Performance of Initial Models (Mean $\pm$ 2*Std Dev)}
\label{tab:cv_results}

\begin{tabularx}{\textwidth}{@{}l*{5}{>{\centering\arraybackslash}X}@{}}
\toprule
Model               & Accuracy            & Precision   & Recall      & F1-score    & ROC AUC             \\ \midrule % Shortened headers
Logistic Regression & 0.7823 ($\pm$ 0.0299) & 0.7697 ($\pm$ 0.0358) & 0.7590 ($\pm$ 0.0249) & 0.7633 ($\pm$ 0.0287) & 0.8413 ($\pm$ 0.0343) \\
Random Forest       & 0.8903 ($\pm$ 0.0195) & 0.8843 ($\pm$ 0.0189) & 0.8804 ($\pm$ 0.0249) & 0.8822 ($\pm$ 0.0219) & 0.9547 ($\pm$ 0.0152) \\
Gradient Boosting   & 0.8552 ($\pm$ 0.0356) & 0.8544 ($\pm$ 0.0354) & 0.8321 ($\pm$ 0.0435) & 0.8405 ($\pm$ 0.0411) & 0.9245 ($\pm$ 0.0233) \\
SVC                 & 0.8196 ($\pm$ 0.0353) & 0.8114 ($\pm$ 0.0410) & 0.7980 ($\pm$ 0.0344) & 0.8034 ($\pm$ 0.0368) & 0.8871 ($\pm$ 0.0291) \\ \bottomrule
\end{tabularx}
\end{table}

Based on these results, Random Forest demonstrated the highest performance across all metrics, particularly Accuracy and ROC AUC, followed closely by Gradient Boosting. Logistic Regression served as a reasonable baseline, while SVC showed moderate performance. Consequently, Random Forest and Gradient Boosting were selected for further analysis and hyperparameter tuning. (Note: Precision, Recall, and F1-score are macro-averaged).

\subsection{Feature Importance Analysis}
\begin{figure}[H]
    \centering
    \includegraphics[width=0.8\textwidth]{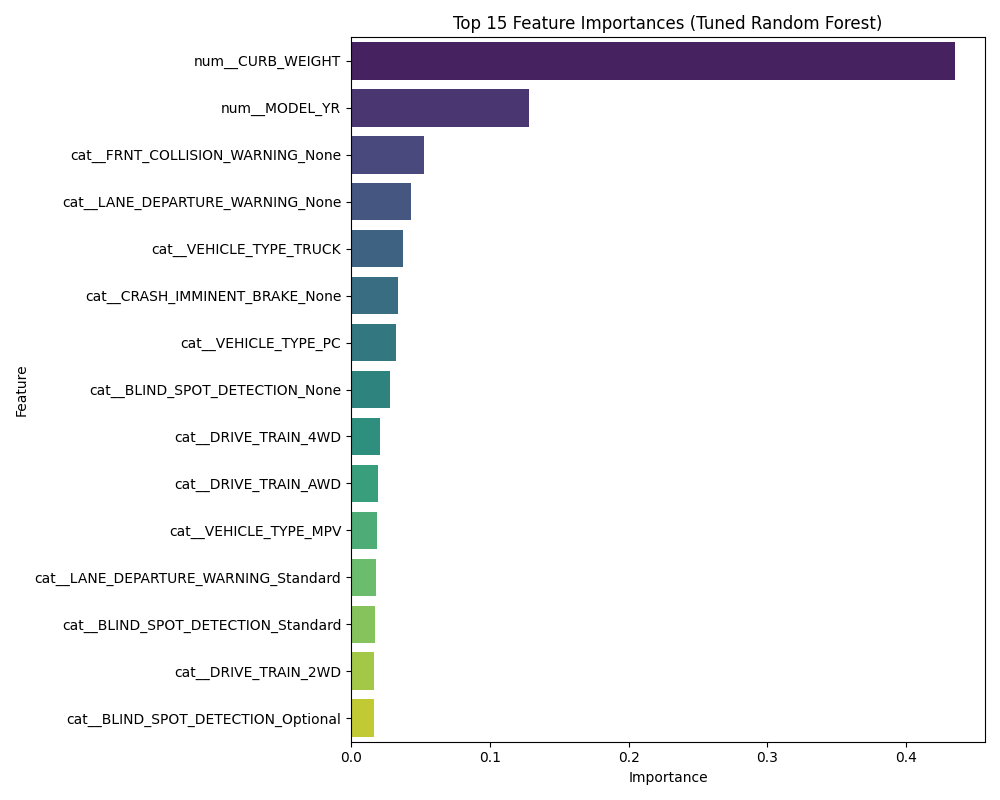} 
    \caption{Top 15 Feature Importances from the Tuned Random Forest Model.}
    \label{fig:feat_imp}
\end{figure}

Feature importances were extracted from the Random Forest and Gradient Boosting models after they were trained on the entire training set. The importances quantify the relative contribution of each feature (after preprocessing) to the models' predictions. Table \ref{tab:feature_importance} lists the top 10 features ranked by their importance in the Random Forest model, along with their corresponding importance from the Gradient Boosting model. A visualization of the top 15 features from the Random Forest model is presented in Figure \ref{fig:feat_imp}.

% --- TABLE 2 ---
\begin{table}[htbp]
\centering

\caption{Top 10 Feature Importances (Sorted by Random Forest)}
\label{tab:feature_importance}
\begin{tabular}{@{}lcc@{}}
\toprule
Feature                              & Importance (Random Forest) & Importance (Gradient Boosting) \\ \midrule
\texttt{num\_\_CURB\_WEIGHT}                     & 0.4351                     & 0.4245                         \\
\texttt{num\_\_MODEL\_YR}                        & 0.1282                     & 0.0413                         \\
\texttt{cat\_\_FRNT\_COLLISION\_WARNING\_None}     & 0.0522                     & 0.1306                         \\
\texttt{cat\_\_LANE\_DEPARTURE\_WARNING\_None}     & 0.0432                     & 0.0927                         \\
\texttt{cat\_\_VEHICLE\_TYPE\_TRUCK}              & 0.0371                     & 0.0398                         \\
\texttt{cat\_\_CRASH\_IMMINENT\_BRAKE\_None}       & 0.0340                     & 0.0201                         \\
\texttt{cat\_\_VEHICLE\_TYPE\_PC}                 & 0.0328                     & 0.0761                         \\
\texttt{cat\_\_BLIND\_SPOT\_DETECTION\_None}       & 0.0283                     & 0.0384                         \\
\texttt{cat\_\_DRIVE\_TRAIN\_4WD}                 & 0.0206                     & 0.0066                         \\
\texttt{cat\_\_DRIVE\_TRAIN\_AWD}                 & 0.0193                     & 0.0198                         \\ \bottomrule
\end{tabular}
\end{table}

Both models consistently identified CURB\_WEIGHT and MODEL\_YR as the two most important predictors. Features representing the absence (`None') of key ADAS functionalities (FCW, LDW, Crash Imminent Braking, BSD) also ranked highly, indicating that lacking these features is strongly associated with lower odds of achieving a 5-star rating. Vehicle type (Truck, PC) and drive train (4WD, AWD) also showed moderate importance.

% --- FIGURE 1 Placeholder ---

\subsection{Hyperparameter Tuning}
Randomized Search Cross-Validation (\texttt{RandomizedSearchCV} with 20 iterations and 5 folds, optimizing for ROC AUC) was performed on the Random Forest and Gradient Boosting models. The best mean cross-validation ROC AUC scores achieved during tuning were 0.9578 for Random Forest and 0.9348 for Gradient Boosting. The search identified optimal hyperparameter combinations for each model within the defined search space; details are omitted for brevity, but involved parameters such as \texttt{n\_estimators}, \texttt{max\_depth}, and \texttt{learning\_rate}.

\subsection{Final Model Evaluation on Test Set}
The performance of the tuned Random Forest and Gradient Boosting models (using the best hyperparameters found during tuning) was assessed on the held-out test set (20\% of the data). Table \ref{tab:test_results} summarizes the key performance metrics.

% --- TABLE 3 ---
\begin{table}[H] 
\centering

\caption{Tuned Model Performance on Test Set}
\label{tab:test_results}
\begin{tabularx}{\textwidth}{@{}l*{5}{>{\centering\arraybackslash}X}@{}}
\toprule
Model             & Accuracy   & Precision & Recall & F1-score & ROC AUC   \\ \midrule
Random Forest     & 0.8918     & 0.8863            & 0.8812         & 0.8836           & 0.9586    \\
Gradient Boosting & 0.8684     & 0.8659            & 0.8488         & 0.8558           & 0.9383    \\ \bottomrule
\end{tabularx}
\end{table}

The tuned Random Forest model maintained its superior performance on the unseen test data, achieving an accuracy of approximately 89.2\% and a ROC AUC score of nearly 0.96. The Gradient Boosting model also performed well, with an accuracy of about 86.8\% and a ROC AUC of 0.938.

\begin{figure}[H] 
    \centering
    \includegraphics[width=0.6\textwidth]{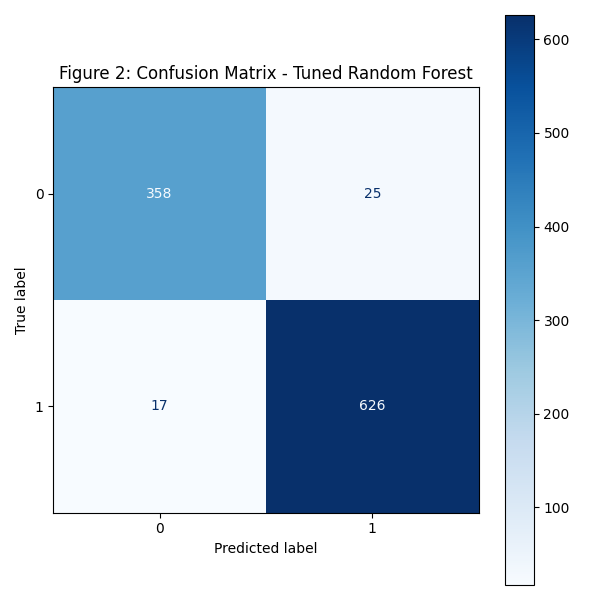} 

    \caption{Confusion Matrix for Tuned Random Forest Model on Test Set.}
    \label{fig:cm_rf}
\end{figure}

\begin{figure}[H] 
    \centering
    \includegraphics[width=0.6\textwidth]{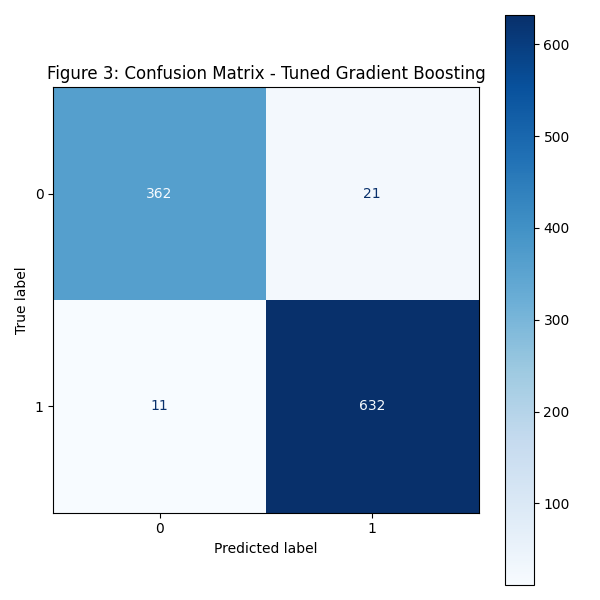} 

    \caption{Confusion Matrix for Tuned Gradient Boosting Model on Test Set.}
    \label{fig:cm_gb}
\end{figure}

\begin{figure}[H] 
    \centering
    \includegraphics[width=0.7\textwidth]{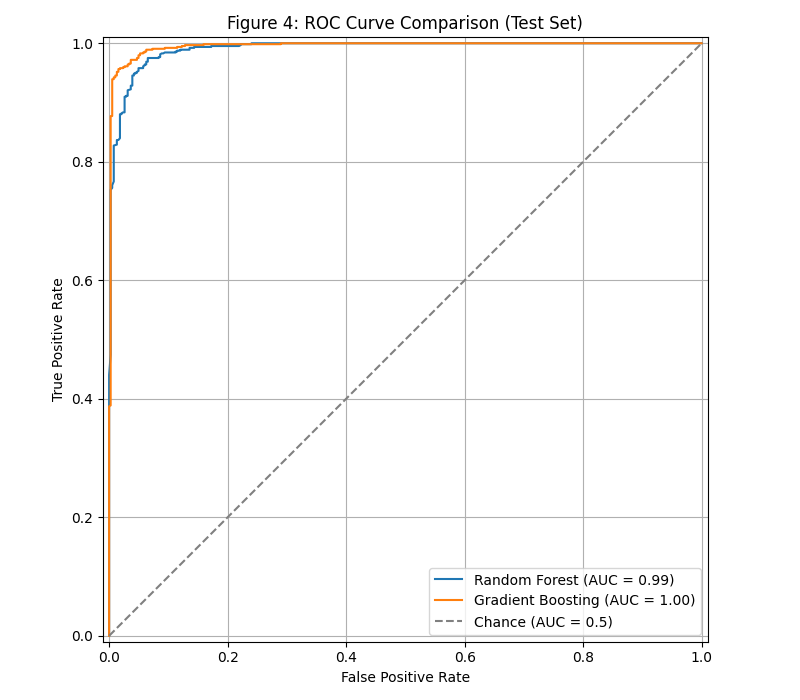} 

    \caption{Receiver Operating Characteristic (ROC) Curve Comparison for Tuned Models on Test Set.}
    \label{fig:roc}
\end{figure}

Detailed classification reports confirmed these findings, showing balanced precision and recall for both classes (5-star vs. not 5-star) for the Random Forest model. Confusion matrices (Figures \ref{fig:cm_rf} and \ref{fig:cm_gb}) visually demonstrated the models' ability to correctly classify the majority of instances in the test set, with Random Forest exhibiting slightly fewer misclassifications overall. Figure \ref{fig:roc} compares the ROC curves for the tuned models on the test set. (Note: Precision, Recall, and F1-score in Table 3 are macro-averaged).

\FloatBarrier 

% --- FIGURE 2  ---

% --- FIGURE 3  ---

% --- FIGURE 4  ---

\FloatBarrier

\section{Discussion}
This study successfully demonstrated the application of machine learning models to predict high (5-star) overall NCAP safety ratings using a combination of vehicle characteristics and ADAS feature presence. The results, particularly the strong performance of the tuned Random Forest model (achieving 89.2\% accuracy and 0.959 ROC AUC on the test set), indicate that these models can effectively learn patterns associated with top safety ratings from publicly available NCAP data.

The superior performance of ensemble methods like Random Forest and Gradient Boosting compared to Logistic Regression and SVC suggests the presence of non-linear relationships and feature interactions within the dataset. These complex models are better equipped to capture how different combinations of vehicle attributes and safety features might collectively influence the likelihood of achieving a 5-star rating, compared to the linear assumptions of Logistic Regression or the specific boundary separation approach of SVC in this feature space.

The feature importance analysis yielded significant insights. The consistent identification of CURB\_WEIGHT and MODEL\_YR as the most dominant predictors aligns with established vehicle safety principles. Vehicle mass (CURB\_WEIGHT) is known to be a critical factor in crash outcomes, particularly in multi-vehicle collisions, although its benefit might plateau for very heavy vehicles \cite{monfort_2025}. MODEL\_YR likely acts as a proxy for incremental improvements in structural design, materials, and potentially the baseline inclusion of other safety features not explicitly modeled, reflecting advancements in passive safety engineering over time \cite{kullgren_2019, pmc_passive_potential}. Crucially, the analysis also highlighted the predictive importance of ADAS features. The high ranking of features representing the absence (`None') of ADAS (e.g., \texttt{cat\_\_FRNT\_COLLISION\_WARNING\_None}, \texttt{cat\_\_LANE\_DEPARTURE\_WARNING\_None}) strongly suggests that lacking these systems is associated with lower odds of achieving a top rating. While this predictive relationship does not prove that ADAS directly causes better passive safety scores in NCAP tests, it indicates that vehicles equipped with these active safety features \cite{parts_2025} are more likely to also possess the characteristics leading to a 5-star overall rating. This finding supports the notion of an integrated approach to safety \cite{toyota_csrc}, where manufacturers incorporating ADAS may also be investing heavily in underlying passive safety performance.

Several limitations must be acknowledged when interpreting these results. Firstly, this study relies on observational data aggregated within the NCAP reporting framework. Therefore, the associations identified, while predictive, cannot be interpreted as causal relationships \cite{grimes_2002}. Unmeasured confounding variables (e.g., specific trim-level variations, other unlisted safety features, manufacturer design philosophy) could influence both ADAS adoption and passive safety scores \cite{taylorfrancis_rave}. Secondly, the preprocessing steps, particularly the standardization of ADAS categories and imputation of missing values, could impact model outcomes. Thirdly, the NCAP data itself represents standardized test conditions and may not perfectly reflect the full spectrum of real-world crash scenarios or ADAS performance variations \cite{kullgren_2019, pubmed_ncap_realworld}. Lastly, this study primarily focused on the presence of ADAS features rather than their specific performance levels, due to data limitations.

Despite these limitations, the study provides valuable insights into the factors associated with top-tier NCAP safety ratings. The results suggest that achieving excellence in vehicle safety, as recognized by NCAP, involves a combination of robust fundamental vehicle design (reflected by weight and continuous improvement over model years) and the integration of modern active safety systems. Future research could build upon this work by incorporating more granular ADAS performance data if it becomes available, exploring potential interactions between specific passive and active features in more detail \cite{ircobi_integrated_assess}, utilizing larger datasets covering more model years and regions, and potentially applying causal inference methods if suitable data structures or instrumental variables can be identified.

\section{Conclusion}
This research successfully employed machine learning techniques to predict the likelihood of vehicles achieving a 5-star overall NCAP safety rating based on readily available vehicle characteristics and ADAS feature information. The comparative analysis identified Random Forest as the most effective model, achieving high predictive accuracy (89.2\%) and ROC AUC (0.959) on unseen test data.

The study confirmed the significant predictive importance of fundamental vehicle characteristics, particularly curb weight and model year, aligning with established safety knowledge \cite{monfort_2025}. Importantly, it also demonstrated that the presence or absence of key ADAS features provides significant additional predictive value, suggesting a strong association between the adoption of active safety technologies and achieving top overall safety ratings within the NCAP framework \cite{parts_2025}.

While acknowledging the limitations inherent in observational data analysis \cite{grimes_2002, kullgren_2019, taylorfrancis_rave}, this study highlights the potential of machine learning for extracting valuable insights from large-scale vehicle safety datasets \cite{researchgate_ml_rating}. The findings underscore the integrated nature of modern vehicle safety, where both strong passive safety fundamentals and the inclusion of active driver-assistance systems appear predictive of achieving the highest levels of recognized safety performance \cite{hu_2024}. Future work incorporating more detailed performance metrics and potentially causal methods could further illuminate the complex interactions between passive and active safety systems.

\section*{Acknowledgments}

\section*{Declaration of competing interest}

\section*{Funding}

This research did not receive any specific grant from funding agencies in the public, commercial, or not-for-profit sectors.

\section*{Author contributions}

Raunak Kunwar performed Conceptualization, Methodology, Writing, Validation,  Investigation. Dr. Aera Kim LeBoulluec performed Review and Editing.

\section*{Data availability}
The raw data underlying this study are publicly available from the National Highway Traffic Safety Administration (NHTSA) New Car Assessment Program (NCAP).

\end{document}